%% file: acl2023.tex
\pdfoutput=1
\documentclass[11pt]{article}

\usepackage{ACL2023}

\usepackage{times}
\usepackage{latexsym}

\usepackage{amssymb}
\usepackage{amsmath}
\usepackage{bm}
\DeclareMathOperator*{\argmax}{arg\,max}

\usepackage{booktabs}

\usepackage[T1]{fontenc}
\usepackage[utf8]{inputenc}

\usepackage{microtype}

\usepackage{inconsolata}

\title{Minimum Bayes' Risk Decoding for System Combination of Grammatical Error Correction Systems}

\author{Vyas Raina \\
ALTA Institute \\
Department of Engineering \\
  University of Cambridge \\
  \texttt{vr313@cam.ac.uk} \\\And
  Mark Gales \\
ALTA Institute \\
Department of Engineering \\
  University of Cambridge \\
  \texttt{mjfg@cam.ac.uk} \\}

\begin{document}
\maketitle
\begin{abstract}

For sequence-to-sequence tasks it is challenging to combine individual system outputs. Further, there is also often a mismatch between the decoding criterion and the one used for assessment. Minimum Bayes' Risk (MBR) decoding can be used to combine system outputs in a manner that encourages better alignment with the final assessment criterion. This paper examines MBR decoding for Grammatical Error Correction (GEC) systems, where performance is usually evaluated in terms of edits and an associated F-score. Hence, we propose a novel MBR loss function directly linked to this form of criterion. Furthermore, an approach to expand the possible set of candidate sentences is described. This builds on a current max-voting combination scheme, as well as individual edit-level selection. Experiments on three popular GEC datasets and with state-of-the-art GEC systems demonstrate the efficacy of the proposed MBR approach. Additionally, the paper highlights how varying reward metrics within the MBR decoding framework can provide control over precision, recall, and the F-score in combined GEC systems.~\footnote{Code available at: \url{https://github.com/rainavyas/mbr_gec}}

% This paper presents a novel application of Minimum Bayes' Risk (MBR) decoding in the context of sequence-to-sequence grammatical error correction (GEC). Considering the unique nature of the GEC task, the paper conceptualizes each output sequence as a set of "edits," facilitating the use of MBR decoding in the edit space. Furthermore, the paper proposes an enhancement to the conventional max-voting technique by including the output sequences derived from it as additional candidates for MBR decoding. Using a greedy MBR decoding algorithm, the paper explores the edit space for an optimal set of edits, leading to substantial performance improvements. Experiments on three popular GEC datasets and with state-of-the-art GEC systems demonstrate the efficacy of the proposed approach. Additionally, the paper highlights how varying reward metrics within the MBR decoding framework can provide control over precision, recall, and F-score, offering further adaptability and control in GEC tasks.~\footnote{Code available after anonymity period.}
\end{abstract}

\input{introduction}
\input{method}
\input{experiments}
\input{conclusions}

\bibliography{custom}
\bibliographystyle{acl_natbib}

\newpage
\input{appendix}

\end{document}

%% file: introduction.tex
\section{Introduction}

Ensembling, the combination of system outputs, is a powerful technique in deep learning, exploiting diverse model capabilities for robust predictions. Though numerous methodologies exist for system combination~\citep{DBLP:journals/corr/abs-2104-02395}, when there is only access to model outputs, many methods are inapplicable and thus the simplest method becomes the averaging of model outputs. However, for sequence-to-sequence (seq2seq) systems, such as summarization, machine translation, and grammatical error correction (GEC), output averaging is less straightforward. A further challenge with seq2seq tasks is the mismatch between the decoding and assessment criteria. \citet{kumar-byrne-2004-minimum} proposed the utilization of Minimum Bayes' Risk (MBR) decoding as a means to select an output that minimizes the theoretical risk according to a designated reward metric. We propose a novel variant of MBR decoding for GEC to allow for system combination and give better alignment with the assessment criteria.

The nature of a GEC task permits the use of MBR decoding within the "edit"-space. Each output sequence can be represented as a set of "edits" required to transform the input sequence into the output. Consequently, the selection of a single output sequence for GEC can be achieved through MBR decoding with a reward function defined on the set of edits, aligned with the edit-based F-score typically used in GEC assessment criteria. Beyond selection, an additional technique known as max-voting~\citep{tarnavskyi-etal-2022-ensembling} can be employed to combine different sets of edits. We propose an enhancement to the performance achieved through max-voting by treating the output sequences obtained from the combination as additional candidates for MBR decoding. Further, with a greedy MBR decoding algorithm, we explore the edit space to identify other candidate edit sets. Through experiments on three popular GEC datasets and use of state of the art GEC systems (Grammarly's GECToR~\citep{DBLP:journals/corr/abs-2005-12592}), we demonstrate that our MBR decoding approach in the edit space consistently leads to significant performance gains. Further, we also show that by selecting different reward metrics as part of the MBR decoding approach we can provide explicit control over precision, recall and the overall F-score used to assess GEC systems.

\section{Related Work}

\noindent\textbf{Grammatical Error Correction}: Early GEC systems using hand-crafted rules~\citep{naber_2003} were replaced by encoder-decoder architectures, using for example Recurrent Neural Networks~\citep{DBLP:journals/corr/ChoMGBSB14}. Today, many state of the art GEC systems use Transformer-based~\citep{NIPS2017_3f5ee243} encoder-decoder architectures to perform the sequence-to-sequence GEC task~\cite{DBLP:journals/corr/abs-2005-00987, DBLP:journals/corr/abs-2010-03260, DBLP:journals/corr/abs-1909-00502, DBLP:journals/corr/abs-2008-02976, stahlberg-kumar-2020-seq2edits}. However, LaserTagger~\citep{malmi-etal-2019-encode}, the PIE model~\citep{DBLP:journals/corr/abs-1910-02893} and Grammarly's GECToR~\citep{DBLP:journals/corr/abs-2005-12592} are all able to achieve competitive performance using a sequence-to-edit structure for the overall sequence-to-sequence task, where a token can be tagged with edit operations. Once a set of tags have been defined, the edit operations can be applied to the input sequence to generate the grammatically correct output sequence. The GECToR system is particularly efficient at inference as it uses a Transformer encoder followed by softmax over linear layers for edit tag prediction, which is significantly faster than standard sequence-to-sequence GEC system decoders. Further, \citet{wu2023chatgpt} demonstrated that GECToR performs better than the most recent generative large language models, e.g. ChatGPT~\citep{DBLP:journals/corr/abs-2005-14165}, which tend to over-correct, compromising on recall performance. Hence this work uses the GECToR model as its base GEC architecture. \newline

\noindent\textbf{System Combination for seqseq systems}: Individual deep learning systems for classification tasks can be combined in many ways: stacking \cite{wolpert1992stacked}, negative correlation learning \cite{liu1999ensemble}, max-voter schemes \cite{ju2018relative, simonyan2014very} or probability averaging \cite{he2016deep, raina_gales_knill_2020, szegedy2015going}. However, for generative language tasks such as GEC, where the output is a sequence of tokens, many traditional ensembling approaches are inapplicable. Sequence-level ensembling approaches, however, can address this by averaging conditional token level probabilities of multiple systems \cite{sennrich2015improving, freitag2017ensemble, malinin2021uncertainty, gec-ensemble-dist}. However, this approach requires identical member architectures as well as access to the output probabilities of the predicted tokens. With the rising trend of limited black box access to large language models (e.g. ChatGPT~\cite{liu2023summary}), system combination methods that only require the generated output sequences have practical benefit.

With access to only the output sequences from individual seq2seq systems, it is challenging to combine them into a single output. For automatic speech recognition, \citet{goelmbr} select a single output using a simple Minimum Bayes' Risk (MBR) decoding approach~\citep{kumar-byrne-2004-minimum}, where the aim is effectively to select the \textit{most average}/representative output sequence. Similarly \citet{manakul2023cued} use MBR to combine sequences for clinical document summarization. The MBR approach has also recently been applied to machine translation~\citep{rosti-etal-2007-combining, rosti-etal-2007-improved, consensus-decoding, freitag-etal-2022-high, muller-sennrich-2021-understanding, zhang2022rmbr}. For GEC systems, \citet{tarnavskyi-etal-2022-ensembling} propose a \textit{max voting} scheme, where only edits predicted by the majority of individual systems are retained. We further improve GEC performance by applying MBR decoding to a sequence selection set augmented with sequences from max voting. We further enrich this selection space with a greedy search over edits.

%% file: method.tex
\section{Output Sequence Combination for GEC}

A Grammatical Error Correction (GEC) system predicts a grammatically correct output sequence $\mathbf y$ from an input sequence, $\mathbf{x}$. With multiple different GEC system output sequence predictions, $\mathcal Y = \{\mathbf y_1,\hdots, \mathbf y_N\}$, for the same input sequence, $\mathbf x$, it is challenging to combine them into a single, best sequence. It is useful to consider the \textit{edit}-space, where a set of edits, $\mathbf e_n(\mathbf x, \mathbf y_n) = \{e_1, \hdots, e_{|\mathbf e_n|}\}$ can be used to represent each predicted output sequence, $\mathbf y_n$~\footnote{Given an input sequence $\mathbf x$ and an output sequence $\mathbf{y}$ it is simple to create an edit set, using tools such as ERRANT~\citep{bryant-etal-2017-automatic}.}. A single edit in the edit set can be defined fully by an input token in $\mathbf{x}$ and an edit operation to apply (insertion, deletion or substitution). This section describes how Minimum Bayes' Risk decoding can be used in the edit-space to combine the different output sequences in $\mathcal{Y}$.

\subsection{MBR decoding for GEC}
MBR decoding aims to select the most representative output sequence, $\mathbf{y}^*\in\mathcal{Y}$. For GEC, we aim to maximise a reward score $\mathcal R$ in the edit-space that encourages better alignment with the final assessment metric,
\begin{equation} \label{eqn:mbr}
    \mathbf{y}^* = \argmax_{{\mathbf{y}}\in\mathcal Y}\left\{ \mathbb E_{p(\tilde{\mathbf{y}}|x)}[\mathcal R(\tilde{\mathbf{e}}(\mathbf{x},\tilde{\mathbf{y}}), \mathbf{e}(\mathbf{x},\mathbf{y}))] \right\},
\end{equation}
where the reward score, $\mathcal R(\tilde{\mathbf{e}}, \mathbf{e})$, views $\tilde{\mathbf{e}}$ as reference edits and $\textbf{e}$ as the hypothesis/predicted edits. In practice, it is difficult to meaningfully estimate the posterior distribution, $p(\tilde{\mathbf{y}}|x)$ for each output sequence. Hence, we consider only similarly performing systems' output sequences, $\mathcal{Y}^{(\text{c})}\in\mathcal{Y}$ to calculate the expectation of the reward and so we approximate each of these sequences as equiprobable,
\begin{equation}\label{eqn:approx}
    \mathbf{y}^* \approx \argmax_{{\mathbf{y}}\in\mathcal Y^{(\text{s})}}\left\{\frac{1}{|\mathcal Y^{\text{(c)}}|}\sum_{\tilde{\mathbf{y}}\in\mathcal Y^{\text{(c)}}} \mathcal R(\tilde{\mathbf{e}}(\mathbf{x},\tilde{\mathbf{y}}), \mathbf{e}(\mathbf{x},\mathbf{y})) \right\},
\end{equation}
where $\mathcal{Y}^{(\text{s})}$ represents the set of possible output sequences we want to select from.

\subsection{MBR decoding with edit voting}\label{sec:voting}

Inspired by \citet{tarnavskyi-etal-2022-ensembling} the different edit sets, $\{\mathbf{e_1},\hdots, \mathbf{e}_N\}$ associated with the different output sequences, can be combined to create a single edit set, $\mathbf{e}^{(m)}$ containing all the individual edits present in at least $m$ of the edit sets (i.e. $m$ \textit{votes}). This new combined edit set represents a new combined output sequence, $\mathbf y^{(m)}$. The MBR decoding approach of Equation \ref{eqn:mbr} can now be applied by simply including the combined sequence in the set of sequences to select from, such that $\mathbf y^{(m)}\in\mathcal{Y}^{(\text{s})}$. Note that the voting scheme can generate a maximum of $N$ different combined sequences, with $\mathbf e^{(1)}$ being the union of all edit sets and $\mathbf e^{(N)}$ the intersection. Hence the selection space of sequences $\mathcal{Y}^{(\text{s})}$ can be made richer with an extra $N$ sequences.

\subsection{Greedy MBR decoding for edit selection}\label{sec:greedy}

Instead of augmenting the selection set $\mathcal{Y}^{(\text{s})}$ with only a few sequences, it is useful to consider all possible edit sets. However, it is computationally infeasible to consider every possible edit set. Hence, this work proposes a practical, greedy method to increase the richness of the selection set. The minimal edit set is arguably the intersection of all edit sets, $\mathbf e^{(N)}$. In contrast the set of possible edits is given by the union set, $\mathbf e^{(1)}$. Hence, we can insert individual edits one by one from the union set to the intersection set. Every new edit insertion into the existing edit set represents a new output sequence $\mathbf y$ (that can be added to $\mathcal{Y}^{(\text{s})}$). However, we only retain the edit insertions that give a new output sequence that increases the MBR expected reward, $\frac{1}{|\mathcal Y^{\text{(c)}}|}\sum_{\tilde{\mathbf{y}}\in\mathcal Y^{\text{(c)}}} \mathcal R(\tilde{\mathbf{e}}(\mathbf{x},\tilde{\mathbf{y}}), \mathbf{e}(\mathbf{x},\mathbf{y}))$ from Equation \ref{eqn:approx}. This way we can efficiently search a richer selection set, $\mathcal{Y}^{(\text{s})}$ of output sequences to find the best combined output sequence $\mathbf{y}^*$.

\subsection{MBR reward score}\label{sec:reward}

Equation \ref{eqn:mbr} uses a reward score $\mathcal R(\tilde{\mathbf{e}}, \mathbf{e})$ to perform MBR decoding. Careful selection of the reward score allows for control over the desired metric to optimise. We can for example aim to combine systems in a manner that encourages better edit \textit{recall},
\begin{equation} \label{eqn:rec}
    \mathcal{R}^{\text{(rec)}}(\tilde{\mathbf{e}}, \mathbf{e}) = \frac{|\tilde{\mathbf{e}}\cap \mathbf{e}|}{|\tilde{\mathbf{e}}|}.
\end{equation}
Conversely, it may be desirable to have a system with high precision,
\begin{equation} \label{eqn:prec}
    \mathcal{R}^{\text{(prec)}}(\tilde{\mathbf{e}}, \mathbf{e}) = \frac{|\tilde{\mathbf{e}}\cap \mathbf{e}|}{|\mathbf{e}|}.
\end{equation}
However, it is usually desirable to have a GEC system with a good combination of precision and recall, as measured by a F-k score,
\begin{equation}\label{eqn:f05}
    \mathcal{R}^{\text{(f\{k\})}}(\tilde{\mathbf{e}}, \mathbf{e}) = \frac{(1+k^2)|\tilde{\mathbf{e}}\cap \mathbf{e}|}{|\tilde{\mathbf{e}}|k + |\mathbf{e}|}.
\end{equation}
As the precision is more important than recall for GEC systems, this work aligns the reward metric with the F0.5 score. The Jaccard Similarity reward metric is also explored as an alternative in Appendix \ref{sec:jac}.

%% file: experiments.tex
\section{Experiments}

\subsection{Experimental setup}

We evaluate performance of the combined systems on three popular grammatical error correction corpora. \textbf{First Certificate in English (FCE)} corpus~\cite{yannakoudakis-etal-2011-new} is a subset of Cambridge Learner Corpus~\cite{openclc_2019} made up of written examinations for general and business English of candidates from 86 different mother tongues, consisting of 2,720 test sentences. \textbf{Building Education Applications 2019 (BEA-19)}~\cite{bryant-etal-2019-bea} offers a test set of 4477 sentences, sourced from essays written by native and non-native English students. \textbf{Conference on Computational Natural Language Learning 2014 (CoNLL-14)}~\cite{ng-etal-2014-conll} test set consists of 1312 sentences sourced from 50 essays written by 25 non-native English speakers. Three different state of the art GECToR models are used as the individual systems to be combined~\footnote{GECToR model Weights: \url{https://github.com/grammarly/gector\#pretrained-models}}. Each system uses a different Transformer encoder (bert (b), roberta (r) or xlnet (x)). Table \ref{tab:base} gives the performance of these individual systems~\footnote{GEC performance for CoNLL and FCE is measured using the ERRANT tool~\citep{bryant-etal-2017-automatic}. Note that CoNLL is often evaluated with a different scorer in other papers. BEA is evaluated using the online submission portal: \url{https://codalab.lisn.upsaclay.fr/competitions/4057}}.

\begin{table}[htb!]
    \centering
    \small
    \begin{tabular}{lccc}
    \toprule
    Model & conll & bea & fce \\ \midrule
        b &  56.15\tiny{$\left(\substack{61.75\\41.19}\right )$} &  65.41\tiny{$\left(\substack{67.33\\58.71}\right )$}& 49.66\tiny{$\left(\substack{54.47\\36.68}\right )$}\\
        r & 56.82\tiny{$\left(\substack{61.99\\42.59}\right )$} & 68.21\tiny{$\left(\substack{70.21\\61.21}\right )$} & 49.86\tiny{$\left(\substack{53.47\\39.28}\right )$}\\
        x & 56.77\tiny{$\left(\substack{61.74\\42.95}\right )$}& 68.00\tiny{$\left(\substack{69.89\\61.36}\right )$}& 50.52\tiny{$\left(\substack{53.49\\41.00}\right )$}\\
         \bottomrule
    \end{tabular}
    \caption{F0.5 and (precision, recall) performance for individual GECToR systems}
    \label{tab:base}
\end{table}

\subsection{Results} \label{sec:results}

MBR decoding (Equation \ref{eqn:approx}) is applied in the edit-space for the three individual GECToR systems' outputs (b,r,x). Here, as the systems have similar performance (equiprobable posterior assumption valid), we let the selection set and the set of sequences to calculate the expected reward be the same  $\mathcal{Y}^{(\text{s})}=\mathcal{Y}^{(\text{c})}=\{b,r,x\}$. Table \ref{tab:mbr} compares the different reward functions, $\mathcal R$, when applying MBR decoding. Selection with precision (Equation \ref{eqn:prec}) and F0.5 (Equation \ref{eqn:f05}) oriented reward metrics give a significant increase in performance over the individual systems in Table \ref{tab:base}. Although the recall reward (Equation \ref{eqn:rec}) does not increase F0.5 performance, it does significantly increase recall performance. This demonstrates that a simple application of MBR decoding can be used to combine individual systems to improve performance and selection of the reward function gives specific control over precision and recall of the combined system.
\begin{table}[htb!]
    \centering
    \small
    \begin{tabular}{lccc}
    \toprule
        Reward & conll & bea & fce \\ \midrule
        $\mathcal{R}^{\text{(rec)}}$ & 55.13\tiny{$\left(\substack{57.76\\46.66}\right )$} & 64.67\tiny{$\left(\substack{64.59\\64.99}\right )$} & 48.74\tiny{$\left(\substack{50.12\\43.90}\right )$}\\
        $\mathcal{R}^{\text{(prec)}}$ & \textbf{59.78}\tiny{$\left(\substack{69.38\\34.48}\right )$} & \textbf{70.87}\tiny{$\left(\substack{75.93\\55.96}\right )$} & \textbf{52.35}\tiny{$\left(\substack{60.13\\34.50}\right )$}\\
        $\mathcal{R}^{\text{(f05)}}$ & 59.71\tiny{$\left(\substack{66.42\\42.53}\right )$} & 69.95\tiny{$\left(\substack{72.95\\60.07}\right )$} & 52.05\tiny{$\left(\substack{56.61\\39.36}\right )$}\\
        \bottomrule
    \end{tabular}
    \caption{MBR with $\mathcal Y^{(\text{c})}= \mathcal Y^{(\text{s})} = \{b,r,x\}$.}
    \label{tab:mbr}
\end{table}

Section \ref{sec:voting} describes how MBR decoding can be applied to systems combined by a voting scheme in the edit space. Table \ref{tab:voting} shows the performance of systems combined with voting, where an individual edit requires $m$ votes (from b,r or x edit system predictions) to be included in the combined edit set, $\mathbf{e}^{(m)}$ to form the single combined sequence $\mathbf{y}^{(m)}$. Note here that $\mathbf{e}^{(1)}$ is the union set and $\mathbf{e}^{(3)}$ is the intersection and so these sequences encourage either a higher recall or precision respectively. Table \ref{tab:mbr-voting} shows the impact of MBR decoding where all the separate voting sets ($\mathbf{y}^{(1)},\mathbf{y}^{(2)},\mathbf{y}^{(3)}$) are included in the selection set, $\mathcal Y^{(\text{s})}=\{b,r,x,\mathbf{y}^{(1)},\mathbf{y}^{(2)},\mathbf{y}^{(3)}\}$. Note that we maintain the same set of sequences for the expected reward calculation, $\mathcal Y^{(\text{s})}=\{b,r,x\}$ to ensure the equiprobable posterior assumption holds~\footnote{Experiments with an alternative set of sequences for $\mathcal Y^{(\text{c})}$ are in Appendix \ref{sec:alt}}. It is evident that a richer selection set allows for even greater improvements in model performance for precision and F0.5 reward MBR decoding.

\begin{table}[htb!]
    \centering
    \small
    \begin{tabular}{lccc}
    \toprule
       System  & conll & bea & fce \\ \midrule
        $\mathbf{y}^{(1)}$ & 47.13\tiny{$\left(\substack{48.02\\43.89}\right )$} & 55.94\tiny{$\left(\substack{55.03\\59.91}\right )$} & 41.76\tiny{$\left(\substack{42.38\\39.43}\right )$}\\
        
        $\mathbf{y}^{(2)}$ & 60.58\tiny{$\left(\substack{68.41\\41.54}\right )$} & 71.82\tiny{$\left(\substack{75.86\\59.22}\right )$} & 52.73\tiny{$\left(\substack{58.16\\38.38}\right )$}\\
        
        $\mathbf{y}^{(3)}$ & 59.60\tiny{$\left(\substack{77.30\\31.10}\right )$} & 72.96\tiny{$\left(\substack{84.32\\47.41}\right )$} & 52.50\tiny{$\left(\substack{67.05\\28.01}\right )$}\\
        \bottomrule
    \end{tabular}
    \caption{Voting combination, $\mathbf{y}^{(m)}$ ($m$ votes).}
    \label{tab:voting}
\end{table}

\begin{table}[htb!]
    \centering
    \small
    \begin{tabular}{lccc}
    \toprule
        Reward & conll & bea & fce \\ \midrule
        $\mathcal{R}^{\text{(rec)}}$ & 53.99\tiny{$\left(\substack{55.99\\47.23}\right )$} & 63.81\tiny{$\left(\substack{63.47\\65.25}\right )$} & 48.18\tiny{$\left(\substack{49.29\\44.20}\right )$}\\
        $\mathcal{R}^{\text{(prec)}}$ & 60.24\tiny{$\left(\substack{76.59\\32.50}\right )$} & \textbf{73.42}\tiny{$\left(\substack{83.40\\49.66}\right )$} & \textbf{53.51}\tiny{$\left(\substack{66.74\\29.85}\right )$}\\
        $\mathcal{R}^{\text{(f05)}}$ & \textbf{60.43}\tiny{$\left(\substack{67.94\\41.90}\right )$} & 70.84\tiny{$\left(\substack{74.48\\59.25}\right )$} & 52.71\tiny{$\left(\substack{57.83\\38.92}\right )$}\\
        \bottomrule
    \end{tabular}
    \caption{MBR with $\mathcal Y^{(\text{c})}=\{b,r,x\}$ and $\mathcal Y^{(\text{s})} = \{b,r,x,\mathbf{y}^{(1)},\mathbf{y}^{(2)},\mathbf{y}^{(3)}\}$.}
    \label{tab:mbr-voting}
\end{table}

Finally, as described in Section \ref{sec:greedy}, MBR decoding can be performed over a richer edit selection space by greedily adding individual edits to the intersection edit set, $\mathbf{e}^{(3)}$ from the union edit set, $\mathbf{e}^{(1)}$. Experiments revealed (Appendix \ref{sec:app-greedy}) that allowing for all edits to be included from the union set can significantly increase the risk of poor insertions, compromising performance. Hence, instead we only consider edits from $\mathbf{e}^{(2)}$ to be added to the intersection set $\mathbf{e}^{(3)}$. Table \ref{tab:greedy} demonstrates that MBR decoding over this richer set of sequences can give better performance (CoNLL) than MBR with voting, but does not always give the best performance (BEA and FCE have better performance in Table \ref{tab:mbr-voting}). This is perhaps because the expected reward over the individual systems (b,r,x) is not necessarily perfectly aligned with the final F0.5 score relative to the true reference edits used in evaluation and thus over-optimisation of the selection set for MBR decoding does not help performance for some datasets.

\begin{table}[htb!]
    \centering
    \small
    \begin{tabular}{lccc}
    \toprule
    Reward & conll & bea & fce \\ \midrule
        $\mathcal{R}^{\text{(rec)}}$ & 61.06\tiny{$\left(\substack{69.50\\41.11}\right )$} &72.20\tiny{$\left(\substack{76.87\\58.08}\right )$} & 52.94\tiny{$\left(\substack{58.87\\37.73}\right )$}\\
        $\mathcal{R}^{\text{(prec)}}$ & 59.65\tiny{$\left(\substack{76.58\\30.79}\right )$} & \textbf{72.76}\tiny{$\left(\substack{84.18\\47.16}\right )$} & 52.62\tiny{$\left(\substack{67.61\\27.89}\right )$}\\
        $\mathcal{R}^{\text{(f05)}}$ & \textbf{61.08}\tiny{$\left(\substack{69.71\\40.85}\right )$} & 72.34\tiny{$\left(\substack{77.16\\57.19}\right )$} & \textbf{53.00}\tiny{$\left(\substack{59.07\\37.57}\right )$}\\
        \bottomrule
    \end{tabular}
    \caption{MBR with $\mathcal Y^{(\text{c})}=\{b,r,x\}$ and greedy search for $\mathcal Y^{(\text{s})}$.}
    \label{tab:greedy}
\end{table}

%% file: conclusions.tex
\section{Conclusions}

The combination of sequence-to-sequence grammatical error correction (GEC) systems is challenging. There is also often a mismatch between the decoding criterion and assessment criterion used for GEC systems. This work demonstrates that a novel Minimum Bayes' Risk (MBR) decoding approach within the edit-space can give an effective system combination method that aligns better with the assessment criteria. We further showed that enhancing the selection space to encompass sequences formulated by max-voting over individual edits can further improve system performance. Moreover, the employment of a greedy search strategy, guided by an MBR reward function, can result in performance gains for the combined system. Crucially, the choice of a reward function in the MBR framework gives users the ability to optimize desired characteristics of the combined GEC system, such as precision, recall or the F-score.

\section{Limitations}

This work explored how MBR decoding can be used to combine individual GEC systems, as well as align the combined system's performance to the edit-based F-score used to assess GEC systems. Experiments were performed with Grammarly's GECToR based systems. It would be useful to extend these experiments to other state of the art GEC systems. Although these other systems are not as efficient as GECToR due to the use of an auto-regressive Transformer decoder (as opposed to GECToR's encoder only structure), it is still meaningful to understand how these systems react to MBR decoding used for system combination. This is particularly relevant as generative large language models are increasingly used for standard natural language tasks.

\section{Ethics Statement}

This work reports on an efficient method to combine individual GEC system outputs in a manner that better aligns with assessment and improve performance. There are no perceived ethical risks associated with this work.

\section{Acknowledgements}

This paper reports on research supported by Cambridge University Press \& Assessment (CUP\&A), a department of The Chancellor, Masters, and Scholars of the University of Cambridge.

%% file: appendix.tex
% APPENDIX:

% 1) Jaccard similarity MBR Reward
% 2) Greedy MBR decoding selection space
% 3) Different Expected Reward Space, $\mathcal{Y}^{(\text{c})}$
% 4) Natural output sequences with MBR -- exps to maximise  
% 5) Upperbound MBR performance - only on FCE

\appendix

\section{Jaccard Similarity as MBR Reward}
\label{sec:jac}

Section \ref{sec:reward} proposes three different reward functions, $\mathcal R$ to guide the MBR decoding process (Equation \ref{eqn:mbr}) to better align with desired assessment criteria. Here, we consider the Jaccard Similarity as an alternative reward function that can combine precision and recall properties,
\begin{equation} \label{eqn:prec}
    \mathcal{R}^{\text{(jac)}}(\tilde{\mathbf{e}}, \mathbf{e}) = \frac{|\tilde{\mathbf{e}}\cap \mathbf{e}|}{|\tilde{\mathbf{e}}\cup \mathbf{e}|}.
\end{equation}
Table \ref{tab:jac} gives the performance of systems combined with MBR decoding using the Jaccard reward function. 

\begin{table}[htb!]
\small
    \centering
    \begin{tabular}{p{1.5cm}ccc}
    \toprule
       $\mathcal Y^{(\text{s})}$  & conll & bea & fce\\ \midrule\midrule
       $\{b,r,x\}$  & 59.22\tiny{$\left(\substack{65.39\\43.01}\right )$} & 69.15\tiny{$\left(\substack{71.55\\60.96}\right )$} & 51.56\tiny{$\left(\substack{55.71\\39.73}\right )$}\\ \midrule
       
       $\{b,r,x,\mathbf{y}^{(1)},$ $\mathbf{y}^{(2)},\mathbf{y}^{(3)}\}$ & 59.21\tiny{$\left(\substack{65.50\\42.76}\right )$} & 69.26\tiny{$\left(\substack{71.87\\60.48}\right )$} & 51.63\tiny{$\left(\substack{55.96\\39.45}\right )$}\\ \midrule
       
       Greedy & 60.94\tiny{$\left(\substack{69.84\\40.37}\right )$} & 72.37\tiny{$\left(\substack{77.29\\57.68}\right )$} & 52.92\tiny{$\left(\substack{59.09\\37.33}\right )$}\\
       \bottomrule
    \end{tabular}
    \caption{GEC system performance with Jaccard Similarity reward function for MBR decoding. In all settings, $\mathcal Y^{(\text{s})} =\mathcal Y^{(\text{c})} = \{b,r,x\}$}
    \label{tab:jac}
\end{table}

Comparing to results in the main experiments (Section \ref{sec:results}), it can be seen that using the Jaccard similarity reward gives similar behaviour but slightly worse performance than the F0.5 reward function used for MBR decoding. This is perhaps expected because both metrics encourage good precision and recall, but the final GEC systems are assessed using the F0.5 score. Hence, the Jaccard similarity reward offers a worse approximation to the final assessment metric than an explicit F0.5 reward in MBR decoding.

\section{Greedy MBR Decoding Selection Space}
\label{sec:app-greedy}

Section \ref{sec:greedy} describes an approach where MBR decoding can be used to greedily search over an edit space between the intersection edit set, $\mathbf{e}^{(3)}$ and the union edit set, $\mathbf{e}^{(1)}$ to find a combined edit set that as per the expected reward in the MBR algorithm should give better performance. Results in the main paper in Table \ref{tab:greedy} search the edit space between the intersection set, $\mathbf{e}^{(3)}$ and $\mathbf{e}^{(2)}$. Table \ref{tab:greedy-all} shows that it is sensible to not continue searching for all edits in the union set, $\mathbf{e}^{(1)}$, as searching the entire space compromises performance. This is perhaps due to the increased noise added into the system by potentially including spurious edits from the union set.
\begin{table}[htb!]
    \centering
    \small
    \begin{tabular}{lccc}
    \toprule
       Reward  &  conll & bea & fce\\ \midrule
        $\mathcal{R}^{\text{(rec)}}$ & 53.22\tiny{$\left(\substack{56.48\\43.24}\right )$} & 63.82\tiny{$\left(\substack{65.34\\58.38}\right )$} & 47.73\tiny{$\left(\substack{50.77\\38.52}\right )$}\\
        $\mathcal{R}^{\text{(prec)}}$ & 59.03\tiny{$\left(\substack{76.58\\30.79}\right )$} & 72.58\tiny{$\left(\substack{84.27\\46.67}\right )$} & 52.56\tiny{$\left(\substack{67.71\\27.73}\right )$}\\
        $\mathcal{R}^{\text{(f05)}}$ & 58.42\tiny{$\left(\substack{67.33\\38.19}\right )$} & 69.66\tiny{$\left(\substack{75.05\\54.11}\right )$} & 50.42\tiny{$\left(\substack{57.33\\34.02}\right )$}\\
        \bottomrule
    \end{tabular}
    \caption{Greedy MBR decoding performance with edit search from intersection edit set, $\mathbf{e}^{(3)}$ to the union edit set, $\mathbf{e}^{(1)}$. By considering all possible edits in the union set, we can reduce performance. Hence in the main paper we limit edits to be between $\mathbf{e}^{(2)}$ and $\mathbf{e}^{(3)}$.}
    \label{tab:greedy-all}
\end{table}

\section{Alternative Expected Reward Set, $\mathcal{Y}^{(\text{c})}$}
\label{sec:alt}

Equation \ref{eqn:mbr} for MBR decoding can be simplified to equation \ref{eqn:approx}, where we make the assumption that every sequence $\mathbf{y}\in\mathcal Y^{(\text{c})}$ used to calculate the expected reward is equiprobable (i.e. the posterior distribution is the same). We justify this assumption in the main paper by considering only similarly performing systems to form the set of sequences over which the expected reward is calculated: $\mathcal Y^{(\text{c})} = \{b,r,x\}$. It is interesting consider a situation where we violate/test this equiprobable posterior assumption by considering different possible sequence sets for $\mathcal Y^{(\text{c})}$ and observing the impact on performance after MBR decoding system combination. Table \ref{tab:alt} reports the performance of MBR decoding with different output sequence sets, $\mathcal Y^{(\text{c})}$ used to calculate the expected reward. In comparison to the equivalent results in the main paper in Table \ref{tab:mbr}, it is evident that a deviation from $\mathcal Y^{(\text{c})} = \{b,r,x\}$ does not compromises performance. This demonstrates that it is possible to diverge from the \textit{similar performing system} constraint to validate the equiprobable posterior assumption to generate good combined systems using MBR decoding.
\begin{table}[htb!]
    \centering
    \small
    \begin{tabular}{lccc}
    \toprule
       Reward  &  conll & bea & fce\\ \midrule
        $\mathcal{R}^{\text{(prec)}}$ & 59.83\tiny{$\left(\substack{69.43\\38.52}\right )$} & 70.81\tiny{$\left(\substack{75.81\\56.02}\right )$} & 52.40\tiny{$\left(\substack{60.15\\34.57}\right )$}\\
        $\mathcal{R}^{\text{(f05)}}$ & 59.96\tiny{$\left(\substack{67.76\\41.07}\right )$} & 70.44\tiny{$\left(\substack{74.21\\58.55}\right )$} & 52.09\tiny{$\left(\substack{57.89\\37.19}\right )$}\\
        \bottomrule
    \end{tabular}
    \caption{Impact of changing the set of sequences, $\mathcal Y^{(\text{c})}$ used to calculate the expected reward when using MBR decoding for system combination. We let $\mathcal Y^{(\text{c})}=\{b,r,x,\mathbf{y}^{(1)},\mathbf{y}^{(2)},\mathbf{y}^{(3)}\}$. In all settings we maintain the same selection set, $\mathcal Y^{(\text{s})} = \{b,r,x\}$.}
    \label{tab:alt}
\end{table}

% \section{Upperbound MBR performance}
% \label{sec:upper}